\documentclass[letterpaper, 10 pt, conference]{ieeeconf}
\IEEEoverridecommandlockouts                              

\usepackage{cite}
\usepackage{amsmath,amssymb,amsfonts}
\usepackage{algorithmic}
\usepackage{graphicx}
\usepackage{textcomp}
\usepackage{xcolor}

\usepackage{algorithm}
\usepackage{algorithmic}
\usepackage{stfloats} 
\usepackage{array}
\usepackage{makecell}

\title{\LARGE \bf
PolyMerge: A Novel Technique aimed at Dynamic HD Map Updates Leveraging Polylines
}

\author{Mohamed Sayed$^{1}$, Stepan Perminov$^{2}$ and Dzmitry Tsetserukou$^{3}$
\thanks{$^{1}$Mohamed Sayed is with the Intelligent Space Robotics Laboratory, Center for Digital Engineering, Skolkovo Institute of Science and Technology, Moscow, Russia
        {\tt\small mohamed.sayed@skoltech.ru}}%
\thanks{$^{2}$Stepan Perminov is with the Intelligent Space Robotics Laboratory, Center for Digital Engineering, Skolkovo Institute of Science and Technology, Moscow, Russia, and with the LLC IntegraNT, Moscow, Russia
        {\tt\small stepan.perminov@skoltech.ru}}%
\thanks{$^{3}$Dzmitry Tsetserukou is with the Intelligent Space Robotics Laboratory, Center for Digital Engineering, Skolkovo Institute of Science and Technology, Moscow, Russia
        {\tt\small d.tsetserukou@skoltech.ru}}%
}

\begin{document}

\maketitle
\thispagestyle{empty}
\pagestyle{empty}

\begin{abstract}

Currently, High-Definition (HD) maps are a prerequisite for the stable operation of autonomous vehicles. Such maps contain information about all static road objects for the vehicle to consider during navigation, such as road edges, road lanes, crosswalks, and etc. To generate such an HD map, current approaches need to process pre-recorded environment data obtained from onboard sensors. However, recording such a dataset often requires a lot of time and effort. In addition, every time actual road environments are changed, a new dataset should be recorded to generate a relevant HD map.

This paper addresses a novel approach that allows to continuously generate or update the HD map using onboard sensor data. When there is no need to pre-record the dataset, updating the HD map can be run in parallel with the main autonomous vehicle navigation pipeline.

The proposed approach utilizes the VectorMapNet framework to generate vector road object instances from a sensor data scan. The PolyMerge technique is aimed to merge new instances into previous ones, mitigating detection errors and, therefore, generating or updating the HD map.

The performance of the algorithm was confirmed by comparison with ground truth on the NuScenes dataset. Experimental results showed that the mean error for different levels of environment complexity was comparable to the VectorMapNet single instance error. 

\end{abstract}


\section{Introduction}
\subsection{Motivation}

In recent years, the development of autonomous vehicles has garnered significant attention as a transformative technology poised to revolutionize transportation. Central to the successful deployment of self-driving cars is the availability of highly accurate and detailed maps for safe operation. These maps, often called High-Definition (HD) maps \cite{rong2020lgsvl}, serve as a critical foundation, enabling vehicles to process surrounding environment and to make informed decisions in real-time. While traditional mapping methods have provided a starting point, the complexity and precision required for autonomous driving necessitate the creation of HD maps using automated techniques. Traditional manual map creation has been replaced by Simultaneous Localization and Mapping (SLAM) algorithms, where a human first manually drives the vehicle and records data from onboard sensors. Data, collected by survey cars or crowdsourcing, is sent to the data center for merging into a 3D map. However, building HD maps faces challenges such as costly and time-consuming data collection and annotation, maintenance difficulty, and large data sizes. Maps can become invalid due to environmental changes, such as snow drifts, road crossings, or road network modifications.

\begin{figure}[t]
    \centerline{\includegraphics[width=0.5\textwidth]{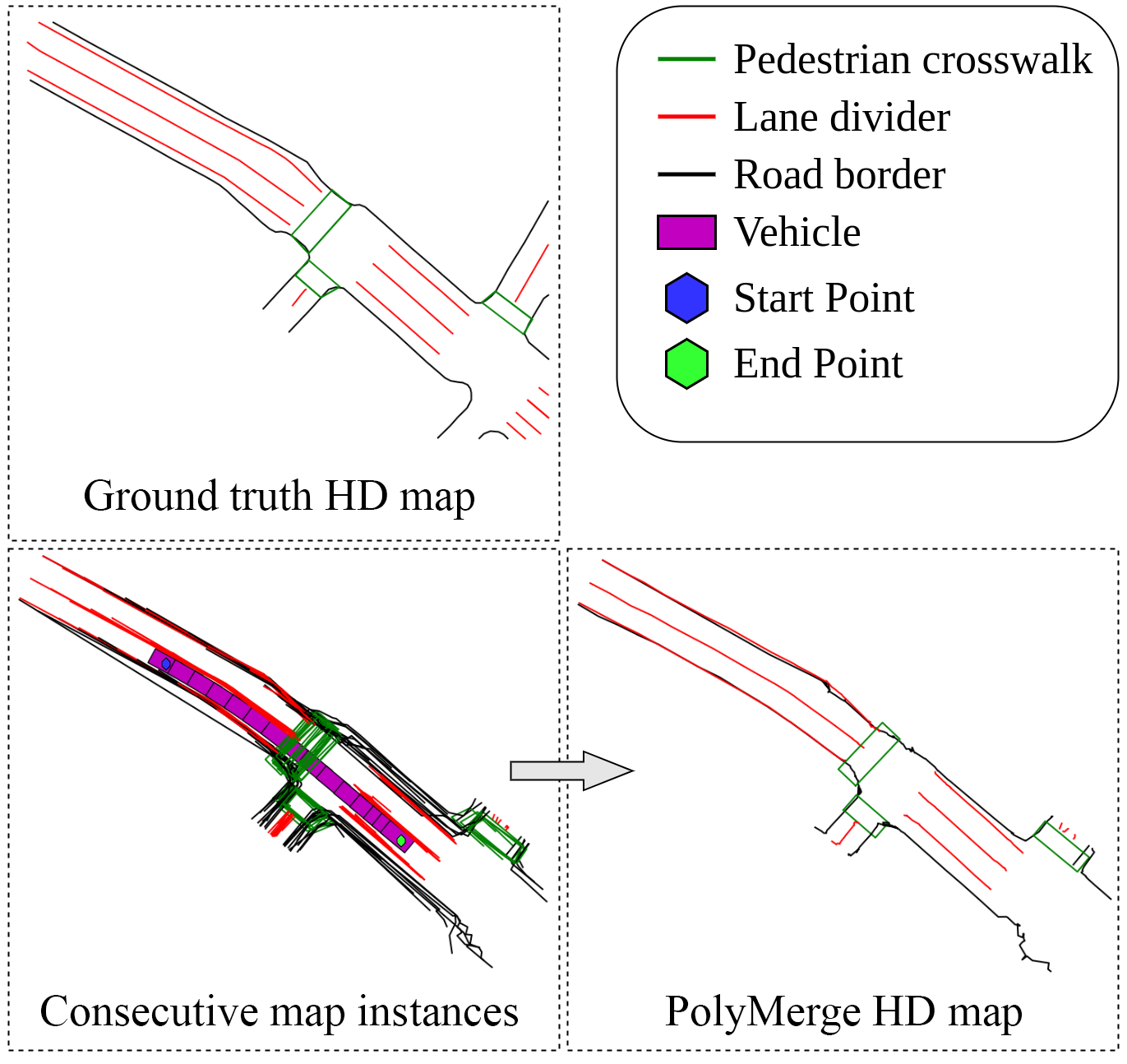}}
    \caption{Ground truth HD map and the PolyMerge HD map, gained from 15 consecutive map instances, NuScenes ``boston-seaport'' map.}
    \label{polymerge_main}
\end{figure}

Recent research, such as Tesla’s Full Self-Driving feature, aims to allow point-to-point navigation without requiring HD map created a priori but it still requires tremendous amounts of training data. Other methods, such as the VectorMapNet \cite{liu2022vectormapnet}, creates a map that can be generated continuously as the vehicle drives through the environment. The VectorMapNet algorithm uses a Convolutional Neural Network (CNN) to extract features from the input image and a Recurrent Neural Network (RNN) to generate the vector map. The CNN is used to learn the patterns of the input image, while the RNN is used to generate the vector map based on the learned features.

The main advantage of the VectorMapNet, which is the focus of this paper, is using polylines as a representation for HD maps.
Polylines provide a geometrically simple and efficient way to represent road shapes and connectivity, allowing for compact storage, streamlined processing, and easier analysis of map data. They offer smooth and continuous representations of road geometry, enabling precise modeling of curved roads, roundabouts, and complex intersections, which is crucial for autonomous vehicles. Additionally, polylines offer flexibility and scalability, allowing for adjustments in HD map detail, accommodating road network changes, and scaling to dynamic environments. Their compatibility and seamless integration with existing map data formats in cartography and GIS make them a versatile choice for various mapping applications.

However, obtained local vector map instances still need to be merged manually, modifying the polyline points, to obtain the global HD map for the autonomous vehicle to navigate.

\subsection{Problem statement}

\begin{figure}[t]    \centerline{\includegraphics[width=0.45\textwidth]{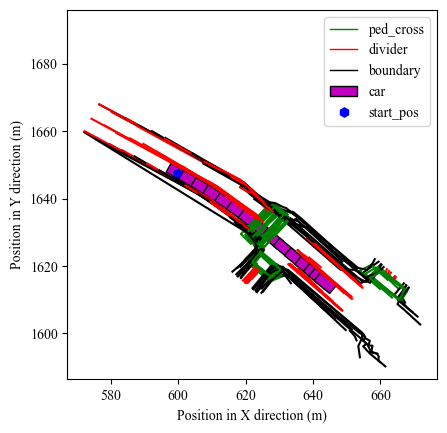}}
    \caption{Concatenated global map of 15 consecutive map instances in bird eye view (BEV) global frame $\mathcal{F}_{W}$, NuScenes ``boston-seaport'' map.}
    \label{unrefined_map}
\end{figure}

This paper addresses the problem of continuous automatic update of HD maps by directly updating the road element polylines, aiming to eliminate the need for manual post-processing. Each time step, autonomous vehicles utilize data from onboard sensors, including LiDAR, cameras, and Global Positioning System (GPS), to generate local HD maps using techniques, such as the VectorMapNet. However, due to imperfections in these techniques, the resulting maps may have intersecting parts that do not align seamlessly with previous maps, as depicted in Fig. \ref{polymerge_main} and Fig. \ref{unrefined_map}. Consequently, multiple polylines with slight shape variations representing the same road element emerge, necessitating the reduction of these polylines into a single merged polyline that encompasses all relevant features. 

\subsection{Related works}

Most recent approaches require a pre-recorded dataset to generate an HD map for an autonomous vehicle.
K.-W. Chiang et al. \cite{chiang2022automated} propose an automated modeling of road networks for HD maps in OpenDRIVE format using point clod data from LiDAR sensors. The algorithm is divided into three phases: lane lines’ extraction from point clouds, modelling lane lines with attributes, and building an OpenDRIVE file.
M. Elhousni et al. \cite{elhousni2020automatic} propose a deep learning based method capable of generating labelled HD maps from raw LiDAR and camera data, pre-collected from a test vehicle.
Y. Zhou et al. \cite{zhou2021automatic} propose an approach based on semantic-particle filter to tackle the automatic lane-level mapping problem. It performs semantic segmentation on 2D front-view images from ego vehicles and explores the lane semantics on a birds-eye-view domain with true topographical projection.

Multiple studies \cite{jo2018simultaneous}, \cite{kim2018crowd}, \cite{kim2021updating} address HD map updates.
K. Jo et al.\cite{jo2018simultaneous} propose an approach that incrementally adds new feature layers during driving and optimizes them using GraphSLAM \cite{grisetti2010tutorial}. The optimized layer is uploaded to a map cloud, where multiple vehicles' layers are combined through data association algorithms. The map cloud updates the integrated layer using a Recursive Least Square (RLS) algorithm whenever new layers are uploaded \cite{simon2006optimal}.
C. Kim et al. \cite{kim2018crowd} propose a crowd-sourced mapping process of the new feature layer for the HD map. Multiple intelligent vehicles are used to acquire new features in the environment to build feature layers for each vehicle using the HD map-based GraphSLAM approach \cite{grisetti2010tutorial}. New feature layers are conveyed to a map cloud through a mobile network system. Finally, crowd-sourced new feature layers are integrated into a new feature layer in a map cloud.
C. Kim et al. \cite{kim2021updating} propose a crowd-sourcing framework to update point cloud maps from environment changes continuously using LiDAR and vehicle communication. While having an initial point cloud map, each vehicle is localized inside it using a hierarchical SLAM approach. The estimated pose is used to detect the differences between the point cloud map and environments, which are defined as map changes that are eventually merged into the point cloud map.
However, none of these studies directly update the final polylines.

Another track is updating the final polylines map by merging the resulting polylines with the main map. While several GIS tools, such as ArcGIS \cite{ArcGIS}, QGIS \cite{QGIS}, Global Mapper \cite{GlobalMapper}, and GRASS GIS \cite{GRASSGIS}, offer functions or capabilities for merging polylines, they still require manual modification of individual points to achieve the desired merging. Moreover, these tools do not address the precise problem of handling multiple lines that overlap with each other or share overlapping sections. Their primary focus lies in merging endpoints through trimming or extending polylines to intersection points or joining already connected lines into a single polyline.

\subsection{Contribution}

To overcome disadvantages of existing map updating techniques and polyline joining tools, this paper introduces the PolyMerge, an automated technique for merging HD maps that specifically focuses on merging polylines, see Fig. \ref{polymerge_main}. The PolyMerge identifies chains of polylines that require merging in a given map instance, taking into account their corresponding labels representing road element types. The technique extends the primary polylines while modifying their overlapping sections with other secondary polylines within the chain. By automating the merging process, the PolyMerge aims to achieve accurate and efficient HD map merging without the necessity of manual point modifications.

\section{Methods}
\begin{figure*}[t]    
    \centerline{\includegraphics{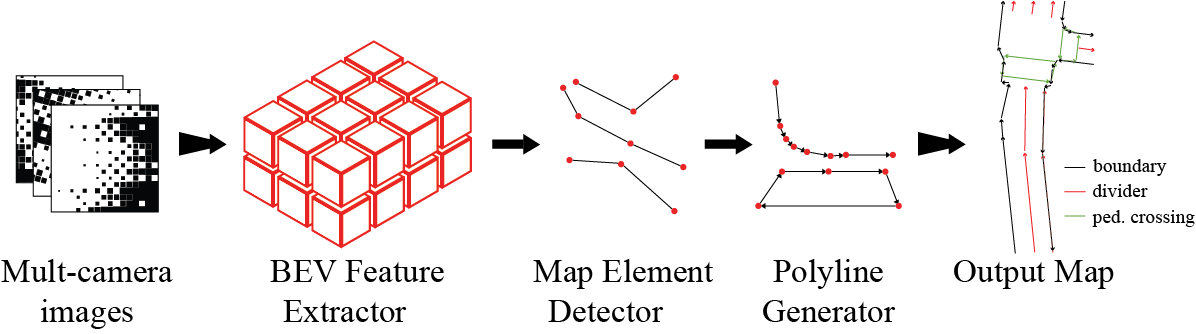}}
    \caption{VectorMapNet system overview.}
    \label{vmn}
    \vspace*{-0.4cm}
\end{figure*}

\subsection{Local HD map construction}

We employed the VectorMapNet to generate the local HD maps due to its utilization of a polyline-based representation rather than a dense collection of semantic pixels. The HD map generation pipeline of the VectorMapNet consists of three essential components as shown in Fig. \ref{vmn}:

\begin{itemize}
    \item A BEV feature extractor for mapping sensor data to a canonical BEV representation;
    \item A scene-level map element detector that identifies and classifies all map elements by predicting element key- points and their class labels;
    \item An object-level polyline generator that generates a sequence of polyline vertices for each detected map element.
\end{itemize}

For training and evaluating the VectorMapNet model, we utilized the NuScenes full dataset (v.1.0)\cite{caesar2020nuscenes}. The NuScenes dataset is a large-scale public dataset for autonomous driving, featuring annotations of 23 object classes with accurate 3D bounding boxes, object-level attributes, and a significant number of camera images, LiDAR sweeps, RADAR sweeps, and object bounding boxes. In addition, to obtain ground truth data of roads, crosswalks, and etc., the NuScenes map expansion pack with 11 semantic layers (crosswalk, side-walk, traffic lights, stop lines, lanes) was implemented. Image data was used for the training and validation of the results. Three types of road elements were considered such as road borders, road dividers, and pedestrian crossings.


\subsection{PolyMerge technique}
A vector map $\mathcal{M}$ is represented by a sparse set of $N_m$ vectorized primitives, specifically polylines $\mathcal{V}^{poly} = \{V_{1}^{poly},...,V_{N_m}^{poly}\}$ in this context, which serve as representations of the map elements, and their class labels $ \mathcal{L} = \{ L_{1},...,L_{N_m}\}$. Each individual polyline $ V_{i}^{poly} = \{\nu_{i,n} \in{\mathbb{R}^2}|n = 1,...,N_{\nu_{i}} \}$ consists of a series of $ N_{\nu_{i}}$ sequentially arranged vertices $ \nu_{i,n}$.

\hfill

\textbf{Method overview.} First, we transform the input map tokens from bird eye view (BEV) representation of the ego frame $\mathcal{F}_{ego}$ to the global world frame $\mathcal{F}_{W}$. Then, we create network graph $G = (\mathcal{V}^{poly},E)$ of similar polylines to be merged, where $E \subseteq \{ \{V_x^{poly}, V_y^{poly}\} |V_x,V_y \in \mathcal{V}^{poly} \}$ is a set of edges connecting the similar polylines. Finally, similar polylines are merged together under the correct label. 

\hfill

\begin{figure*}
    \centering
    \includegraphics{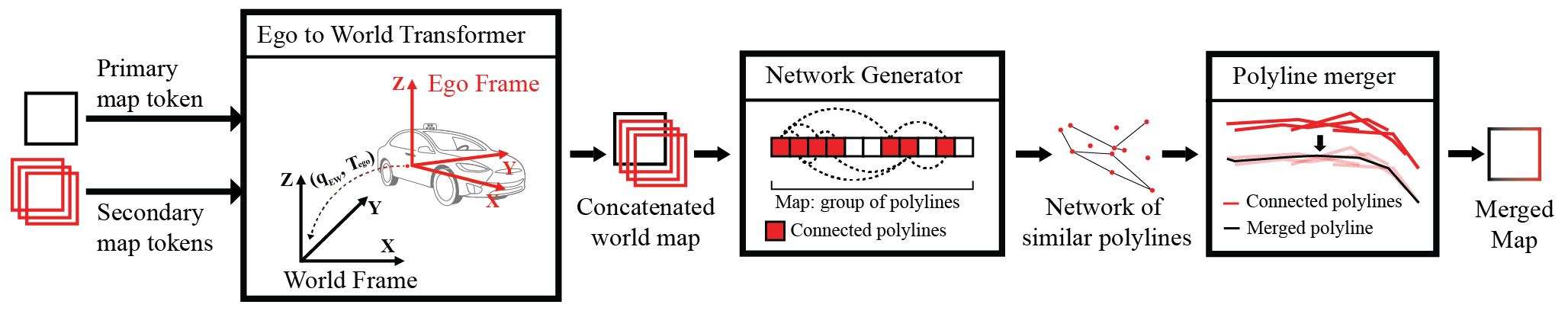}
    \vspace*{-0.2cm}
    \caption{PolyMerge overview}
    \label{PolyMerge_Overview}
\end{figure*}

Correspondingly, the PolyMerge employs three parts to produce the merged map as shown in Fig. \ref{PolyMerge_Overview}: Ego to world frame transformer; Network generator that identifies similar polylines; Polyline merging tool that combines the similar polylines into one merged polyline.

\hfill

\subsubsection{Ego to World transformer} We use simple frame transformations to rotate and then translate each polyline's vertices $\nu_{i,n,_{ego}}$ given in ego frame $\mathcal{F}_{ego}$ to a global frame $\mathcal{F}_{W}$:
\begin{equation}
    \nu_{i,n_{W}} = q \ \nu_{i,n_{ego}} \ q^{-1} + T_{ego} \ ,
\end{equation}

\noindent where $q$ is the rotation quaternion vector from  $\mathcal{F}_{ego}$ to $\mathcal{F}_{W}$, $q \ \nu \ q^{-1}$ is a Hamilton product \cite{goldman2011understanding}, and $T_{ego}$ is the translation vector of the ego frame to the world frame.

It has to be noted that a main map $ \mathcal{M}_m$ has to be defined as the one to merge the other $N_s$ secondary maps $ \mathcal{M}_{s} = \{ \mathcal{M}_{s,_1},...,{M}_{s,_{N_s}}\}$ into. 
Then,  a new map $ \mathcal{M}_{conc}$ is created by concatenating the transformed maps while adding another label  $ \mathcal{L}m = \{ Lm,_{1},...,Lm,_{N_m}\}$ that distinguish main map polylines, where $ \{ Lm,_i = 1 \ if\  V_i^{poly} \in \mathcal{M}_m, \ else\  Lm,_i = 0 \} $.


\subsubsection{Network Generator} Given the concatenated map $ \mathcal{M}_{conc}$ contacting all the maps' polylines in world frame, the goal of network generator is to determine the subsets of polylines that need to be merged together. 

The first step involves iterating through all the polylines belonging to the secondary maps $\mathcal{M}_S$, after which each polyline is examined for its proximity to all other polylines in the concatenated map $\mathcal{M}_{conc}$. To decide if two polylines are close or not,  Euclidean distances are calculated between all points in one polyline and their corresponding projections onto the other polyline. If any of the points from both polylines fall within an acceptable distance from the other polyline (below a defined threshold $Th_{prox}$), indicating close proximity, the polylines are considered suitable for merging if they possess similar labels. The chosen acceptable distance prevents the merging of road elements separated by a road in between and accounts for a margin of error in both elements. It can be adjusted for various road configurations. Finally, using polylines made it easier to distinguish map element types even if they overlap. 

\begin{figure}[b]
    \centering
    \includegraphics[width=0.35\textwidth]{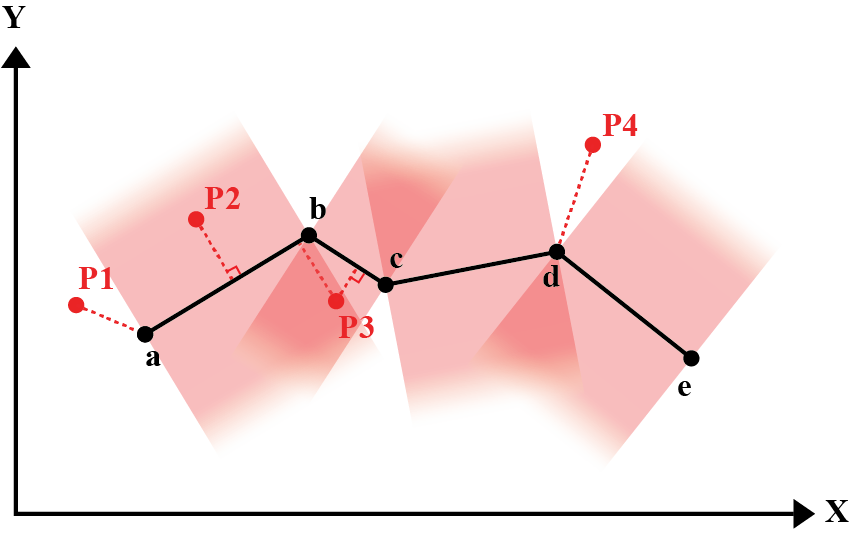}
    \vspace*{-0.2cm}
    \caption{Example of in-line projection.}
    \label{projection}
    \vspace*{-0.3cm}
\end{figure} 

A crucial aspect of this part is determining the projection of a given point onto a polyline. This is accomplished by identifying the nearest line segment of the polyline to the point, which is determined by calculating the minimum distance between the point and its in-line projection. If a direct projection onto the line is not possible, the closest edge serves as the projection, as depicted in Fig. \ref{projection}. The in-line projection $D$ of point $A$ onto line segment $BC$, is calculated as following:
\begin{equation}
    \Vec{D} = \Vec{B} + max(0,min(1,\frac{\overrightarrow{BA} \cdot \overrightarrow{BC}}{\overrightarrow{BC} \cdot \overrightarrow{BC}})) * \overrightarrow{BC}
\end{equation}

These nearest line segment edges are subsequently utilized to position the point within the merged polyline. The algorithm of Network Generator is shown in Alg. \ref{NetworkGenerator}.

 \begin{algorithm}[H]
 \caption{Network Generator algorithm}
 \begin{algorithmic}[1]
 \renewcommand{\algorithmicrequire}{\textbf{Input:}}
 \renewcommand{\algorithmicensure}{\textbf{Output:}}
 \REQUIRE $ \mathcal{M}_{conc}$
 \ENSURE  $G = (\mathcal{V}^{poly},E)$
  \STATE Create empty Network G
  \STATE Define proximity threshold $Th_{prox}$
  \FOR {$V_i^{poly}$ in $\mathcal{M}_{merged} [Lm_i=0]$}
  \FOR {$V_j^{poly}$ in $\mathcal{M}_{merged}$}
  \STATE $check =$ polyline\_merge\_check($V_i^{poly}$,$V_j^{poly}$,$Th$)
  \IF {($check = True$)}
  \STATE G.add\_edge($V_i^{poly}$, $V_j^{poly}$)
  \ENDIF
  \ENDFOR
  \ENDFOR
 \RETURN $G$
 \end{algorithmic}
 \label{NetworkGenerator}
 \end{algorithm}  

\vspace{0.5 cm}
The function ``polyline\_merge\_check'' compares the distance between the polylines' points and their projections to the given threshold as explained, then verifies the labels assigned to the two polylines. If the distance criterion is satisfied and the labels are compatible, the function returns a boolean value of true, indicating that the polylines can be merged. Distance computation details are provided in Section \ref{sec:experimental_results}.

\hfill

\begin{figure}[t]
    \centering
    \includegraphics{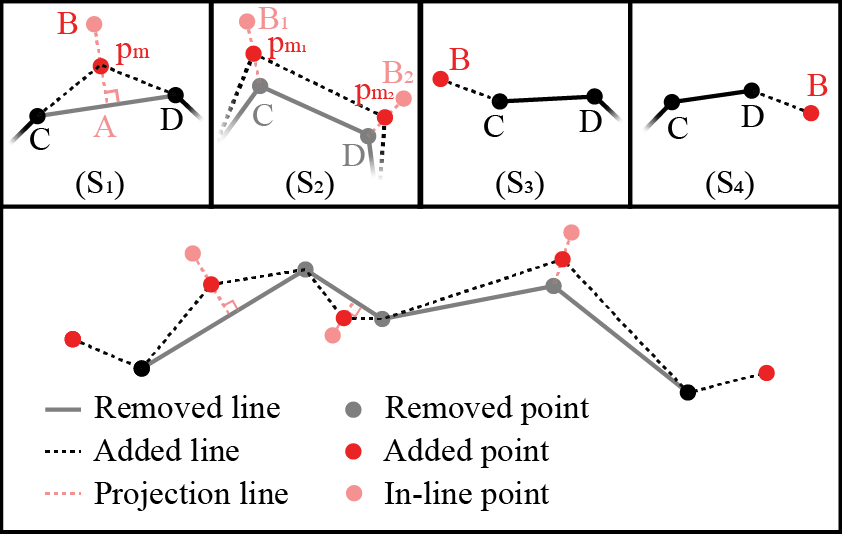}
    \caption{Different scenarios for point to polyline projection.}
    \label{4_scenarios}
    \vspace*{0.3cm}
\end{figure} 

\subsubsection{Polyline Merging}
Following the computation of the network graph $G = (\mathcal{V}^{poly},E)$, the PolyMerge proceeds by iterating through the connected graph edges that represent the polylines to be merged. First, it defines the main map polyline $V_i,_{\mathcal{M}}$, which serves as the base onto which the remaining polylines will be merged. Then, one by one, it merges the remaining polylines onto the base polyline. To merge polyline $V_A^{poly}$ onto $V_B^{poly}$, we deal with $V_A^{poly}$ point by point. We could narrow the possible scenarios for merging point $A \in V_A^{poly}$, having its in-line projection point $B$ onto $V_B^{poly}$, into the following four scenarios:
\begin{itemize}
    \item First scenario: Point $B$ falls inside a line segment $ \overline{CD} \subset V_B^{poly}$. In this case, a mid point $P_m = \frac{A+B}{2}$ is inserted onto $V_B^{poly}$ with an index between points $C$ and $D$ indices.
    \item Second scenario: $B$ equals either point $C$ or point $D$. In this case, $P_m$ is also calculated but replaces $C$ or $D$ in $V_B^{poly}$, depending if it equals $C$ or $D$.
    \item Third scenario: $B$ equals $C$ and $C$ is an edge point of $V_B^{poly}$. In this case, $B$ is appended into $V_B^{poly}$ before $C$, as the new starting edge.
    \item Fourth scenario: $B$ equals $D$ and $D$ is an edge point of $V_B^{poly}$. In this case, $B$ is appended into $V_B^{poly}$ after $D$, as the new ending edge.
\end{itemize}

The four scenarios are explained in Fig. \ref{4_scenarios}. $V_B^{poly}$ is then updated with every iteration. 

\begin{figure}[b]
    \centering
    \includegraphics[width=0.45\textwidth]{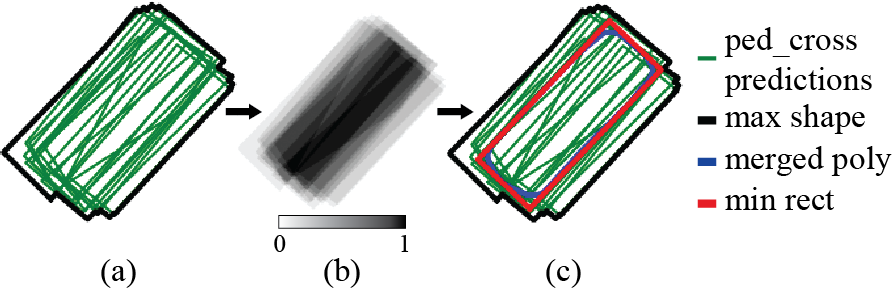}
    \caption{Example of pedestrian crossing merging by the PolyMerge technique.}
    \label{cross}
    \vspace*{-0.3cm}
\end{figure}

However, specific road elements such as pedestrian crossings, always come in quadrilateral closed shapes, and thus we used a better method for their merge. First, similar quadrilaterals are rasterized onto a grid, and an empty rectangle is created around their union shape. The cells within the rectangle are filled based on the number of quadrilaterals covering them. This rectangle represents the probability of polygon coverage, ranging from 0.0 to 1.0. A modified Gaussian blur function is applied using full convolution to avoid edge distortions. Finally, the average result is extracted by selecting a quantile threshold $Th_{cov}$, such as 0.1 for larger coverage or 0.95 for smaller coverage. A quantile threshold of 0.5 represents the median, or an ``average'' coverage area. Finally, we use a rotated minimum bounding rectangle to represent the pedestrian crossing.
Fig. \ref{cross} explains the described merging method and compares results with the case of merging them in the same way as other road elements. 


\section{Experimental Results}
\label{sec:experimental_results}
\begin{figure}[t]
    \centering
    \includegraphics[width=0.4\textwidth]{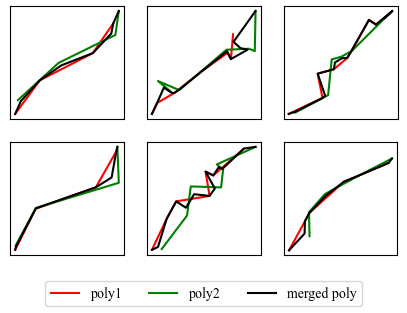}
    \caption{Examples of polyline merging by the PolyMerge technique.}
    \label{samples}
    \vspace*{-0.3cm}
\end{figure}

Different polyline compositions have been tested for merging individually as shown in Fig. \ref{samples}. It shows how the PolyMerge technique effectively captures and averages distances between two polylines, considering the selection of points at the start and end.

Demonstration of merging multiple map instances by the Polymerge technique is provided in Fig. \ref{polymerge_main}. It is shown that polylines effectively encode detailed geometries and direction information of map elements, reflecting real-world structures and explicit directions. The merging algorithm is able to successfully preserve detailed geometry while considering all registered map instances. However, some parts (e.g. road border at lower right corner) were not accurately estimated by the VectorMapNet method, which resulted in chaotic merged polyline.

\begin{figure}[t]
    \centerline{\includegraphics[width=0.51\textwidth]{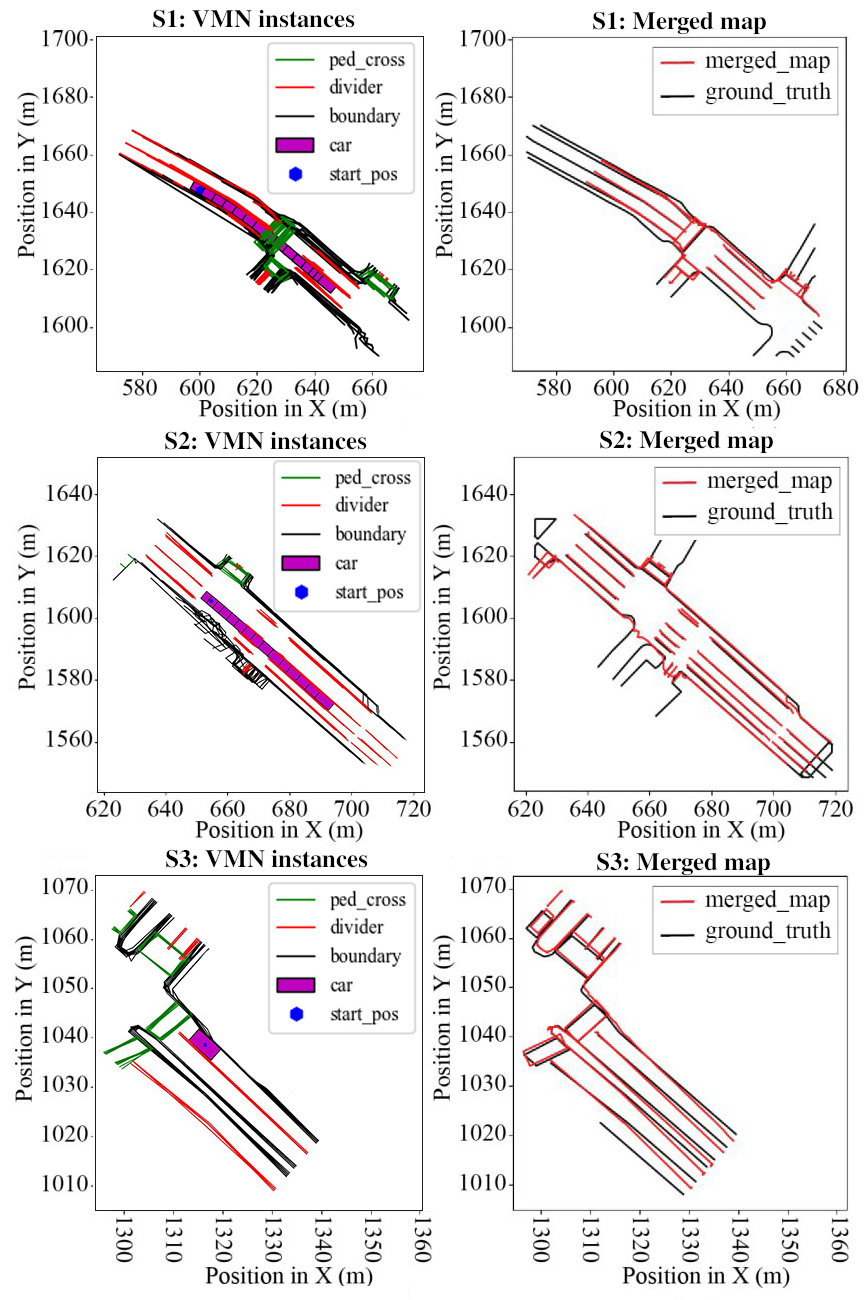}}
    \vspace*{-2mm}
    \caption{Comparison of HD map gained by the PolyMerge algorithm from 15 consecutive map instances in three different scenarios (S1, S2 and S3) with the ground truth HD map in bird eye view (BEV) global frame $\mathcal{F}_{W}$, NuScenes ``boston-seaport'' map.}
    \label{overlap_rb}
\end{figure}

In order to validate the performance of the developed PolyMerge technique, we prepared three experimental scenarios representing data from the NuScenes dataset of different road sections. For each scenario, 15 local HD map instances were obtained by the VectorMapNet (VMN) and then processed by the PolyMerge technique to generate the merged HD map, see Fig. \ref{overlap_rb}. Both VMN instances and the merged map were compared with corresponding Ground Truth data using the ``polyline\_merge\_check'' method for detecting corresponding polylines from Alg.\ref{NetworkGenerator}. 
We used the Partial Curve Method (PCM) \cite{witowski2012parameter} to measure the difference between each pair of corresponding polylines. PCM normalizes vectors and matches the area of a subset between the two curves, allowing identification of the optimal section of the ground truth poly that corresponds to a short estimated poly. Additionally, we calculated the discrete Frechet distance (DF) \cite{eiter1994computing}, which measures the similarity between curves while considering the location and ordering of points along the curve. DF represents the length of the shortest leash that allows traversing both curves, similar to a man walking a dog on a leash without backtracking.
For the experiment, a proximity threshold $Th_{prox}$ of 1 m, a quantile threshold $Th_{cov}$ of 0.5, and the VMN model prediction confidence of 0.8 were applied.


Experimental results are provided in Table \ref{difference}, where a comparison of average metric values measured on polylines from 15 VMN local HD map instances with metric values of the corresponding merged HD maps is provided. As shown, the PolyMerge generally preserves mean DF and PCM values with slight variations compared to VMN instances. DF distances show closer alignment to the ground truth for pedestrian crossings and boundaries, reducing maximum error. However, dividers show increased errors due to VMN inaccuracies and unaccounted extra dividers merging with overlapping instances. Higher PCM values suggest a need for additional smoothing step due to increased zigzag movements. To mitigate these issues, improving VMN accuracy is essential. Currently reliant on camera images, future enhancements may include integrating lidar scans for more precise road element detection and expanding class categories beyond pedestrian crossings, dividers, and boundaries.


\begin{table}[t]
\caption{Processed Results of HD Map Experiments}
\vspace*{-5mm}
\label{difference}
\begin{center}
\noindent
\setlength\tabcolsep{4.3pt} 
\begin{tabular}{ |p{0.07cm}|p{0.9cm}|p{0.65cm}|p{0.65cm}|p{0.65cm}|p{0.65cm}|p{0.65cm}|p{0.65cm}|p{0.65cm}| }
\hline
\multicolumn{3}{|c|}{\textbf{Map}}&\multicolumn{3}{|c|}{\textbf{PCM}} &\multicolumn{3}{|c|}{\textbf{DF distance (m)}}\\
\cline{4-9} 
\multicolumn{3}{|c|}{\textbf{}}&\textbf{\textit{$ped$}}&\textbf{\textit{$div$}}&\textbf{\textit{$bou$}}&\textbf{\textit{$ped$}}&\textbf{\textit{$div$}}&\textbf{\textit{{$bou$}}}\\
\hline
 & & Mean &1.06 & \textbf{0.20} &  \textbf{0.26} & 1.30 & \textbf{0.49} & 1.65\\
 &VMN & Min & \textbf{0.24} & \textbf{0.02} & \textbf{0.05} & \textbf{0.16} & \textbf{0.09} & 0.38\\
 &instances & Max & 1.93 & \textbf{1.29} & \textbf{1.80} & 4.95 & 1.16 & 5.93\\
\multicolumn{1}{|c|}{\textbf{S1}} && STD & 0.37 & \textbf{0.27} & \textbf{0.31} & 1.36 & \textbf{0.28} & 0.94\\
 
\cline{2-9} 
& & Mean & \textbf{0.63} & 0.77 & 1.71 & \textbf{0.91} & 0.59 & \textbf{1.21}\\
&Merged & Min & 0.56 & 0.28 & 1.51 & 0.39 & 0.28 & \textbf{0.29}\\
 &map& Max & \textbf{0.73} & 1.55 & 2.11 & \textbf{1.42} & \textbf{0.89} & \textbf{1.91}\\
 && STD & \textbf{0.09} & 0.60 & 0.35 & \textbf{0.52} & 0.34 & \textbf{0.83}\\

\hline\hline
 & & Mean &1.33 & 0.43 &  \textbf{0.23} & \textbf{0.88} & \textbf{0.27} & 2.3\\
 & VMN& Min & \textbf{0.24} & \textbf{0.02} & \textbf{0.02} & \textbf{0.29} & \textbf{0.08} & \textbf{0.42}\\
 &instances& Max & 2.42 & 4.08 & \textbf{0.70} & 1.48 & 0.51 & 6.68\\
\textbf{S2} && STD & \textbf{1.09} & 0.91 & 0.25 & 0.60 & \textbf{0.13} & 2.26\\
 
\cline{2-9} 
& & Mean & \textbf{1.26} & \textbf{0.18} & 0.53& 1.25 & 0.35& \textbf{1.02}\\
&Merged & Min & 0.48 & 0.03 & 0.19 & 1.05 & 0.18 & 0.52\\
 &map& Max & \textbf{2.04} & \textbf{0.55} & 0.75 & \textbf{1.45} & 0.57 & \textbf{1.60}\\
 && STD & 1.10 & \textbf{0.17} & \textbf{0.24} & \textbf{0.28} & \textbf{0.13} & \textbf{0.41}\\

\hline\hline
 & & Mean &1.10 & \textbf{0.08} &  \textbf{0.43} & \textbf{0.76} & \textbf{0.67} & \textbf{1.21}\\
 & VMN& Min & 0.40 & \textbf{0.03} & \textbf{0.04} & \textbf{0.53} & \textbf{0.11} & \textbf{0.67}\\
 &instances& Max & 2.26 & \textbf{0.20} & \textbf{1.27} & 1.19 & 1.86 & \textbf{1.71}\\
\textbf{S3} && STD & 0.59 & \textbf{0.05} & \textbf{0.50} & 0.14 & \textbf{0.56} & 0.30\\
 
\cline{2-9} 
& & Mean & \textbf{0.22} & 0.58 & 0.93& 1.03 & 0.86 & 1.43\\
&Merged & Min & \textbf{0.16} & 0.04 & 0.27 & 0.97 & 0.18 & 1.23\\
 &map& Max & \textbf{0.33} & 1.19 & 2.18 & \textbf{1.09} & \textbf{1.65} & \textbf{1.71}\\
 && STD & \textbf{0.09} & 0.53 & 2.08 & \textbf{0.09} & 0.70 & \textbf{0.25}\\

\hline
\end{tabular}
\end{center}
\end{table}


\section{Conclusion}
We introduce the PolyMerge, a novel technique for dynamic HD map updates. In contrast to existing methods, the PolyMerge directly merges polylines of similar road elements using the local vector map generated. By constructing a network graph of similar polylines and projecting them onto each other, an average representation is obtained keeping the offset with ground truth mostly equal to the used map instances with a mean of 1.06 m DF distance for pedestrian crossings, 0.60 m for road dividers, and 1.22 for road boundaries compared to 0.98 m, 0.48, and 1.72 m in the used VMN map instances. Our experiments demonstrate the ease of implementation and effectiveness of the PolyMerge, resulting in a comprehensive map that closely resembles the ground truth. 



\bibliographystyle{IEEEtran}


\end{document}